\documentclass[runningheads]{llncs}
\usepackage[pagebackref,breaklinks,colorlinks]{hyperref}
\usepackage{graphicx}
\usepackage[small]{caption}
\usepackage{graphicx}
\usepackage{amsmath}
\usepackage{amssymb}
\usepackage{booktabs}
\usepackage{algorithm}
\usepackage{algorithmic}
\usepackage[switch]{lineno}

\usepackage{color,wrapfig}
\usepackage{tabularx}
\usepackage{multirow}
\usepackage{booktabs}
\usepackage{makecell}
\usepackage{enumitem}
\setitemize{noitemsep,topsep=0pt,parsep=0pt,partopsep=0pt}
\usepackage{caption}
\usepackage{wrapfig}
\usepackage[capitalize]{cleveref}
\crefname{section}{Sec.}{Secs.}
\Crefname{section}{Section}{Sections}
\Crefname{table}{Table}{Tables}
\crefname{table}{Tab.}{Tabs.}

\begin{document}
\title{Self-Prompt SAM: Medical Image Segmentation via Automatic Prompt SAM Adaptation}

% \author{Paper ID: 1669}
% \institute{}
%
\titlerunning{Self-Prompt SAM}
% If the paper title is too long for the running head, you can set
% an abbreviated paper title here
%
% \author{First Author\inst{1}\orcidID{0000-1111-2222-3333} \and
% Second Author\inst{2,3}\orcidID{1111-2222-3333-4444} \and
% Third Author\inst{3}\orcidID{2222--3333-4444-5555}}
% %
% \authorrunning{F. Author et al.}
% % First names are abbreviated in the running head.
% % If there are more than two authors, 'et al.' is used.
% %
% \institute{Princeton University, Princeton NJ 08544, USA \and
% Springer Heidelberg, Tiergartenstr. 17, 69121 Heidelberg, Germany
% \email{lncs@springer.com}\\
% \url{http://www.springer.com/gp/computer-science/lncs} \and
% ABC Institute, Rupert-Karls-University Heidelberg, Heidelberg, Germany\\
% \email{\{abc,lncs\}@uni-heidelberg.de}}
%

\author{Bin Xie\inst{1} \and
Hao Tang\inst{2} \and 
Dawen Cai\inst{3} \and 
Yan Yan\inst{4} \and 
Gady Agam\inst{1}}

% % TODO FINAL: Replace with an abbreviated list of authors.
\authorrunning{B.~Xie et al.}
% % First names are abbreviated in the running head.
% % If there are more than two authors, 'et al.' is used.

% % TODO FINAL: Replace with your institution list.
% \institute{Princeton University, Princeton NJ 08544, USA \and
% Springer Heidelberg, Tiergartenstr.~17, 69121 Heidelberg, Germany
% \email{lncs@springer.com}\\
% \url{http://www.springer.com/gp/computer-science/lncs} \and
% ABC Institute, Rupert-Karls-University Heidelberg, Heidelberg, Germany\\
% \email{\{abc,lncs\}@uni-heidelberg.de}}
\institute{Department of Computer Science, Illinois Institute of Technology, USA \\ 
 \and
School of Computer Science, Peking University, China\\
 \and
Department of Cell and Developmental Biology, University of Michigan, USA\\
 \and
Department of Computer Science, University of Illinois Chicago, USA \\ 
\email{\small{bxie9@hawk.iit.edu, haotang@pku.edu.cn,\\
dwcai@umich.edu, yyan55@uic.edu, agam@iit.edu}}
}

\maketitle              % typeset the header of the contribution
%
% \vspace{-0.4cm}
\begin{abstract}
Segment Anything Model~(SAM) has demonstrated impressive zero-shot performance and brought a range of unexplored capabilities to natural image segmentation tasks. However, as a very important branch of image segmentation, the performance of SAM remains uncertain when applied to medical image segmentation due to the significant differences between natural images and medical images. Meanwhile, it is harsh to meet the SAM's requirements of extra prompts provided, such as points or boxes to specify medical regions.
In this paper, we propose a novel self-prompt SAM adaptation framework for medical image segmentation, named Self-Prompt-SAM. We design a multi-scale prompt generator combined with the image encoder in SAM to generate auxiliary masks. Then, we use the auxiliary masks to generate bounding boxes as box prompts and use Distance Transform to select the most central points as point prompts.
Meanwhile, we design a 3D depth-fused adapter~(DfusedAdapter) and inject the DFusedAdapter into each transformer in the image encoder and mask decoder to enable pre-trained 2D SAM models to extract 3D information and adapt to 3D medical images. 
Extensive experiments demonstrate that our method achieves state-of-the-art performance and outperforms nnUNet by 2.3\% on AMOS2022~\cite{ji2022amos}, 1.6\% on ACDC~\cite{bernard2018deep} and 0.5\% on Synapse~\cite{landman2015miccai} datasets.

\keywords{Medical Image Segmentation  \and Automatic Prompt \and SAM}

\end{abstract}
\vspace{-0.6cm}
\section{Introduction}
\label{sec:intro}
    \vspace{-0.2cm}
    
The purpose of medical image segmentation is to utilize medical images to segment specific anatomical structures, including organs, lesions, and tissues, which can aid in many clinical applications. Deep learning methods~\cite{ronneberger2015u,akkus2017deep,chen2021transunet,zhou2021nnformer,isensee2019automated,xie2024ms} have made remarkable and numerous progress in the field of medical image segmentation in the past few years. However, existing deep learning models are often tailored, which have a strong inductive bias and limit their capacity.
% for specific tasks.

The rise of foundation models~\cite{brown2020language,openai2023gpt4} that are trained on large and diverse datasets has revolutionized artificial intelligence. Benefiting from their remarkable zero-shot and few-shot generalization abilities, a wide range of downstream tasks that adapt a pre-trained model to specific tasks~\cite{houlsby2019parameter,pan2022st,hu2021lora,yang2023aim} achieve remarkable progress, not like traditional methods of training task-specific models from scratch. Recently, SAM~\cite{kirillov2023segment}, pre-trained over 1~billion masks on 11~million natural images, has been proposed as a visual foundation model for prompt-driven image segmentation and has gained huge attention due to its impressive zero-shot performance.
Based on its strong capabilities in natural image segmentation, can SAM still maintain strong performance when applied to medical image segmentation, though significant differences between natural images and medical images? 

% \begin{figure*}[t!]
% \centering
%     %\vspace{-0.6cm}
% \includegraphics[width=0.9\textwidth]{fig/prove_prompt.pdf}
%       \caption{Experiments to prove which combination of points, boxes, and masks is the best prompt.} 
% % \textcolor{red}{[include more detailed descriptions.]}}
%   \vspace{-0.6cm}
% \label{fig:prompt}
% \end{figure*}

\begin{figure*}[t!]
\centering
  %\vspace{-0.6cm}
\includegraphics[width=1\textwidth]{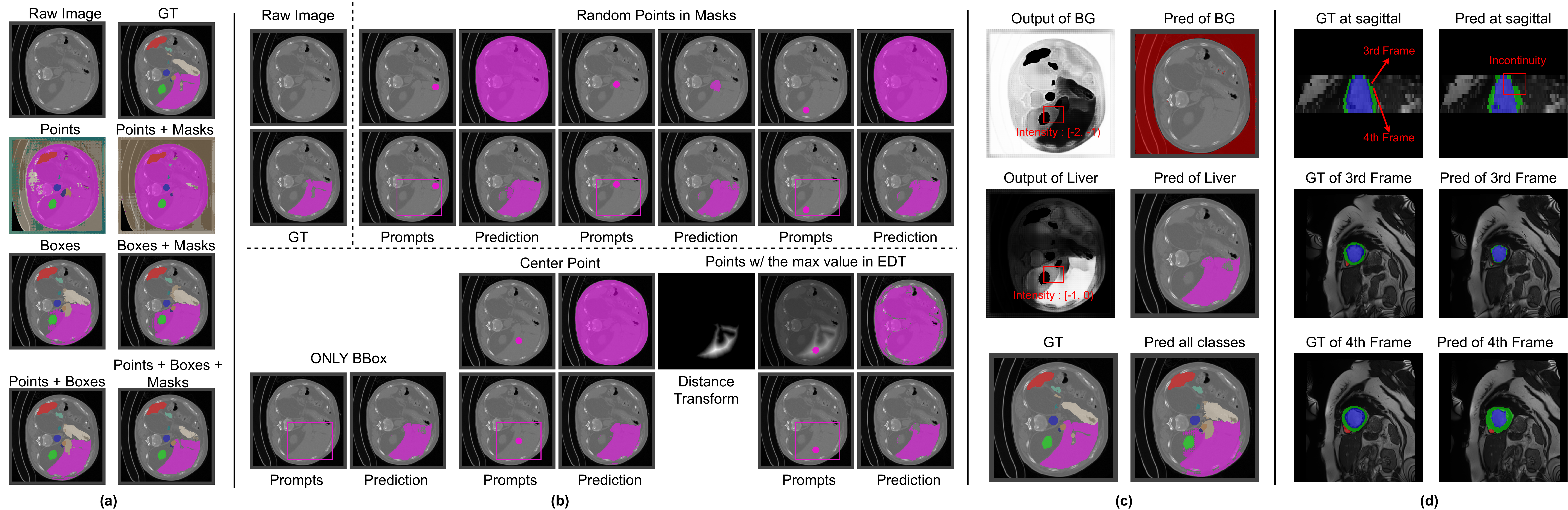}
      \caption{(a) Predictions of different combinations of points, boxes, and masks from ground truth. (b) Experiments for different methods to select points. (c) The intensity distribution and prediction of the output of SAM. (d) Incontinuity at the depth dimension.} 
% \textcolor{red}{[include more detailed descriptions.]}}
  \vspace{-0.6cm}
\label{fig:point-prompt}
\end{figure*}

It is infeasible to apply directly. We present the intrinsic issues when directly applying SAM to medical image segmentation as follows. i) The first issue is that SAM needs extra prompts when segmenting specific regions. It is harsh to expect all users to have medical knowledge to provide points in specific regions or frame out specific regions. 
% Figure~\ref{fig:prompt}(a) shows different prompts, such as points, boxes, and masks. When a user without any medical background is given a raw image, it is hard to identify every class of organs, such as the liver~(purple region). Therefore, how to solve the harsh requirements is the first key issue when applying SAM to medical image segmentation. 
ii) The second issue is that SAM does not have the functionality to predict semantic information for predicted binary masks. SAM only predicts one binary mask for each prompt without any semantic labels.
% , which is more like for instance segmentation methods. 
However, medical images usually have multiple labels and each label has semantic information. 
% Our expectation of SAM is to predict each class given raw images, like traditional medical image segmentation. 
iii) The third issue is which combination of prompt ways achieves the best performance when proper prompts are provided. SAM needs users to provide mandatory prompts that could be either points and boxes or both, and an optional prompt, masks. 
iv) The fourth issue is that directly applying SAM to medical image segmentation tasks without any modification or fine-tuning does not always obtain good performance when proper prompts are provided. Many works~\cite{deng2023segment,hu2023sam,zhou2023can,mohapatra2023sam} have demonstrated that SAM is imperfect or even fails when some situations occur, such as weak boundaries, low-contrast, and irregular shape, which is consistent with other investigations~\cite{ji2023sam,ji2023segment}. Figure~\ref{fig:point-prompt}(a) illustrates the results using different prompts. Even if the prompts are generated by the ground truth, the results are very bad, especially by point prompts.

To solve the requirements of extra prompts, we propose a multi-scale prompt generator~(MSPGenerator) combined with the image encoder that employs the Vision Transformer~(ViT)~\cite{dosovitskiy2020image} pre-trained with masked auto-encoder~\cite{he2022masked} as the backbone to generate auxiliary multi-class masks that can produce bounding boxes as box prompts and points as point prompts instead of manual prompts. 
% We found the image encoder employs the Vision Transformer~(ViT)~\cite{dosovitskiy2020image} pre-trained with masked auto-encoder~\cite{he2022masked} as the backbone. Benefiting from ViT's strong representation capabilities, 
% the image encoder extracts essential features of the images with a series of transformer blocks. Consequently, 
% we can extract multiple levels of feature maps, serving as inputs for our MSPGenerator, which learns to generate auxiliary masks. 
% MSPGenerator is built via a hierarchical structure with convolutional layers to learn different levels and different scale information by gradually upsampling the spatial dimensions to the same size as the ground truth. 
% Afterwards, auxiliary masks would be utilized for prompts, such as bounding boxes of auxiliary masks as box prompts or points in the foreground as point prompts. 
To solve the functionality to provide semantic information for predicted binary masks, we utilize the generated auxiliary multi-class masks, which can be encoded to one-hot binary masks whose number is the same number of classes. Then, we utilize the location in the channel dimension of the one-hot binary masks to represent the semantic labels. 

% Sometimes, there are binary masks may not have any foreground. We assign the values of both a box prompt and a point prompt to zero. 
% Therefore, each one-hot binary mask specifically is responsible for a certain semantic label by the location or index of the channel dimension in the one-hot binary masks.
% , which can produce box prompts and point prompts. 
% Sometimes, there are binary masks may not have any foreground. In this case, we assign the values of both a box prompt and a point prompt to zero. In this way, we utilize the location in the channel dimension of the one-hot binary masks to represent the semantic labels.  

After acquiring the prompts, we investigate the optimal prompt method for medical image segmentation. In Figure~\ref{fig:point-prompt}(a), we found the most robust prompt way is by bounding boxes with proper points that should select the point farthest from the boundary of each object, which means that the point is as central as possible shown in Figure~\ref{fig:point-prompt}(a). Therefore, we adopt the Euclidean distance transform to calculate the distance from the boundaries and obtain the candidate points based on the auxiliary masks.

The final part is to explore how to adapt the original SAM works from 2D natural image segmentation to 3D medical image segmentation. Directly applying SAM to medical image segmentation tasks does not always obtain good performance when proper prompts are provided. 
% Many works~\cite{deng2023segment,hu2023sam,zhou2023can,mohapatra2023sam,roy2023sam,wang2023sam,he2023accuracy} have demonstrated that SAM is imperfect or even fails when some situations occur, such as weak boundaries, low-contrast, and smaller and irregular shape, which is consistent with other investigations~\cite{ji2023sam,ji2023segment}. Figure~\ref{fig:prompt}(b) illustrates results by different prompts. Even if the prompts are generated by the ground truth, the results are very bad. 
Therefore, fine-tuning SAM for medical image segmentation tasks is the main direction. However, fine-tuning the large model of SAM consumes huge computational resources. This problem can be solved by parameter-efficient fine-tuning~\cite{houlsby2019parameter,hu2021lora,yang2023aim}, such as inserting trainable lightweight adapters~\cite{houlsby2019parameter} that prove its feasibility on SAM by the works~\cite {xie2024masksam,ma2023segment,wu2023medical,li2023auto,gong20233dsam}, lightly modifying large models, and freezing the rest of the structures to fine-tune large models efficiently. The question of how to appropriately modify the structure has become the most important issue. To maximize the utilization of the capabilities of SAM, 
% we need to maintain as much structure as possible. To adapt SAM from 2D natural image segmentation to 3D medical image segmentation, we need to modify the whole structure. 
our criteria are to keep all structures, freeze all weights, and only add blocks into SAM to adapt. In this way, we retain the zero-shot capabilities of SAM and adapt SAM to medical image segmentation. Meanwhile, we design several additional blocks to do adaptation.

\begin{figure*}[t]
\centering
    %\vspace{-0.6cm}
\includegraphics[width=0.9\textwidth]{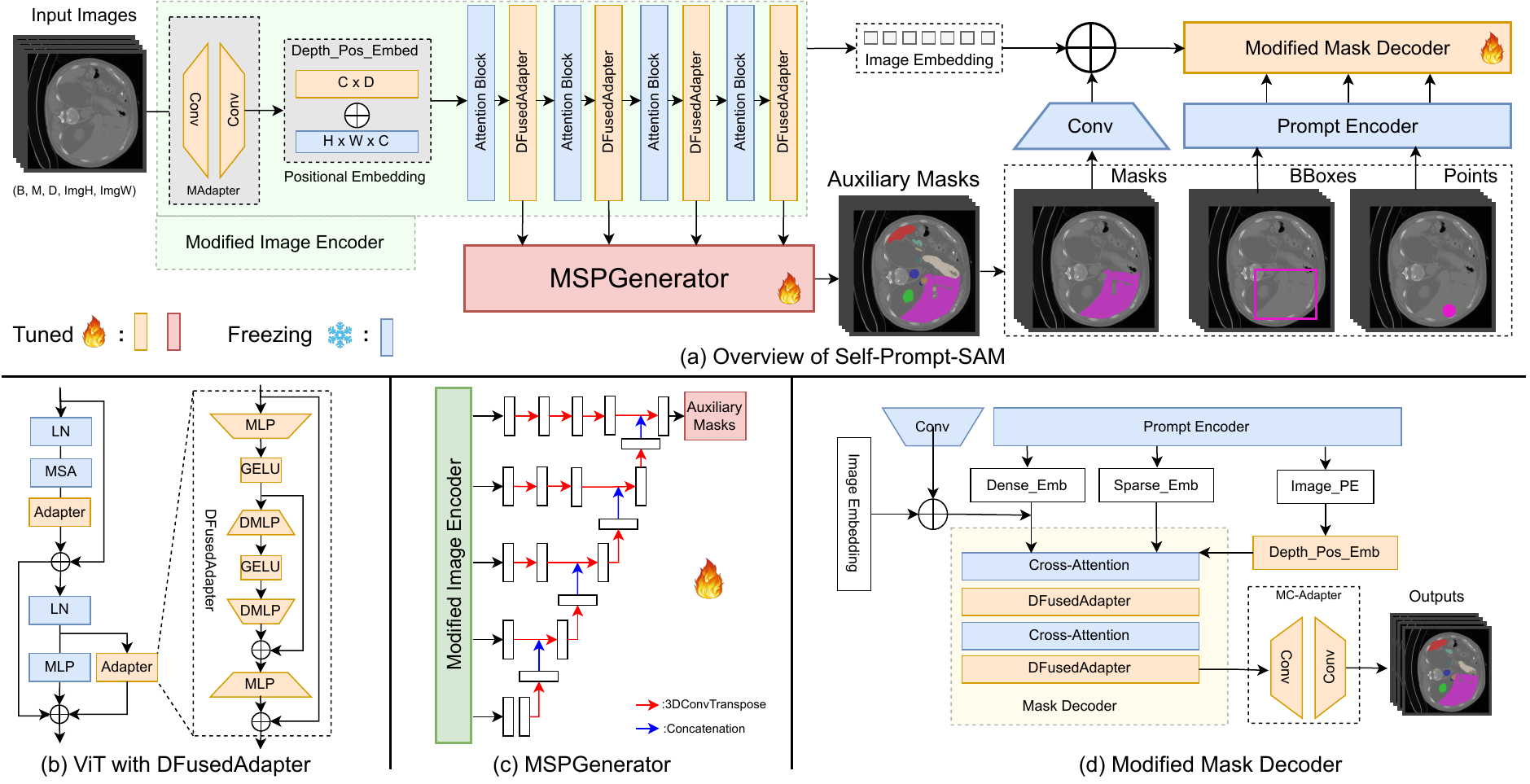}
      \caption{The overview architecture of the proposed  Self-Prompt-SAM.} 
% \textcolor{red}{[include more detailed descriptions.]}}
  \vspace{-0.8cm}
\label{fig:arch}
\end{figure*}

In summary, our contributions to this paper are as follows:
\textbf{(i)} We propose a novel self-prompt SAM~(Self-Prompt-SAM) framework for medical image segmentation. To the best of our knowledge, the proposed Self-Prompt-SAM is the first SAM-based image segmentation framework without any prompts provided;
\textbf{(ii)} We propose a novel multi-scale hierarchical prompt generator~(MSPGenerator) that utilizes multiple levels of feature maps from the image encoder to generate auxiliary masks for the prompts. Through massive experiments, we found that the best prompt way is to combine bounding boxes, points~(use Euclidean distance transform to generate candidate points), and masks;
\textbf{(iii)} We design a depth fused adapter~(DFusedAdapter) to enable pre-trained 2D SAM models to extract 3D information.
\textbf{(iv)}We conduct extensive experiments on three challenging AMOS2022~\cite{ji2022amos}, ACDC~\cite{bernard2018deep} and Synapse \cite{landman2015miccai} datasets. The results demonstrate that Self-Prompt-SAM achieves state-of-the-art performance. 

\vspace{-0.4cm}
\section{Methodology}
\label{sec:method}
\vspace{-0.2cm}

Figure~\ref{fig:arch} illustrates our Self-Prompt-SAM, including a modified image encoder, a designed MSPGenerator, the prompt encoder, and a modified mask decoder.

\vspace{-0.4cm}
\subsection{Proposed DFusedAdapter}
\vspace{-0.2cm}
Since most medical images have an extra depth dimension compared to 2D natural images, it is inevitable to lose 3D spatial information and cause spatial incontinuity in depth dimension if SAM is applied directly to a sequence of 2D frames shown in Figure~\ref{fig:point-prompt}(d). There exists an incontinuity between the 3\textit{rd} frame and the 4\textit{th} frame at the depth dimension when we use SAM without any operation of the depth information. To solve the incontinuity in depth dimension, we design the DFusedAdapter shown in Figure~\ref{fig:arch}(b) with the ability to explore depth information by adding an invert-bottleneck structure consisting of two FC layers processing in depth dimension with an activation layer in the middle of the original adapter with a skip connection. 
% The DFusedAdapter can be expressed as $\text{DFusedAdapter}(\textbf{\textit{X}}) =$ 
% \begin{equation}
%     \textbf{\textit{X}} + (\sigma(\textbf{\textit{X}}\cdot \textbf{\textit{W}}_{dn}) + \sigma (\sigma(\textbf{\textit{X}}\cdot \textbf{\textit{W}}_{dn}) \cdot \textbf{\textit{W}}_{Dup}) \cdot \textbf{\textit{W}}_{Ddn}) \cdot \textbf{\textit{W}}_{up},
% \label{equa:DFusedAdapter}
% \end{equation}
% where $\sigma$ denotes the activation function, $\textbf{\textit{W}}_{dn} {\in} \mathbb{R}^{C\times \frac{C}{4}}$ and $\textbf{\textit{W}}_{up} {\in} \mathbb{R}^{\frac{C}{4}\times C}$ denote the linear down- and up-projection layer processing on the channel dimension respectively, $\textbf{\textit{W}}_{Dup} {\in} \mathbb{R}^{D\times 4D}$ and $\textbf{\textit{W}}_{Ddn} {\in} \mathbb{R}^{{4D}\times D}$ denote the linear up- and down-projection layer processing on the depth dimension respectively. 
% In this way, our model can learn the extra-depth information. 
Inspired by AIM~\cite{yang2023aim}, we introduce an adapter after multi-head self-attention and in parallel to the MLP layer for each transformer block. Thus, our model can learn the extra-depth information.

\vspace{-0.4cm}
\subsection{Modified Image Encoder}
\vspace{-0.2cm}
The modified image encoder is illustrated in Figure~\ref{fig:arch}(a). 
% Given images $X {\in} \mathbb{R}^{B\times M{\times} D\times imgH{\times} imgW}$ with a batch size of $B$, $M$ number of modalities, and a spatial resolution of $D{\times} imgH{\times} imgW$. Our goal is to predict the corresponding pixel-wise segmentation with size ${B\times N{\times} D{\times} imgH{\times} imgW}$, where $N$ is the number of classes of a segmentation task. 
Firstly, the given images with varied modalities will be adapted and reshaped to 3 channels by our designed MAdapter. Then, the adapted images will be fed to the image encoder, which consists of a patch embed block, a positional embedding block that adds a learnable depth positional embedding block to learn extra depth information, and a series of transformer blocks that we insert our designed DFusedAdapter after the MSA and in parallel to the MLP layer to adapt SAM to medical image segmentation and learn extra depth information. 
% After the image encoder, the image embedding is obtained. 
Meanwhile, we extract several feature maps for the MSPGenerator. 

\noindent \textbf{Proposed MAdapter.} 
SAM works on natural images with 3 channels for RGB while medical images have varied modalities. To solve the issue of how to adapt varied modalities to RGB channels, we design an invert-bottleneck architecture built via a sequence of convolutional layers, named MAdapter, at the very beginning of the image encoder, which can learn the adaption during fine-tuning. 
% Moreover, SAM directly upsamples images with any size to a size of $1024\times1024$ as the input of the image encoder. For pixel-level tasks, medical segmentation tasks are sensitive to the intensity of each pixel, especially the boundaries of objects. If we directly upsample images to $1024\times1024$, it is inevitable to lose information and produce ambiguity. 
% Therefore, we designed an invert-bottleneck architecture built via a sequence of convolutional and transposed convolutional layers to learn the adaption from the varied modalities with any size to 3 channels with a size of $1024\times1024$, named MAdapter.

\vspace{-0.4cm}
\subsection{Proposed MSPGenerator}
\vspace{-0.2cm}
% Since SAM needs to provide extra prompts, such as points or boxes, when specific objects need to be segmented, many works of SAM-based medical image segmentation need to provide manual prompts during training and inference. It is not always feasible since we can not expect all users to have medical knowledge to provide proper prompts. The rest of the works for adapting SAM to medical image segmentation usually abandon the prompt encoder or mask decoder to avoid the requirements. However, it is not advisable since it would
% destroy the consistent system of SAM. Meanwhile, it is not wise to abandon the strong prompt encoder and mask decoder, which are trained via large-scale datasets and lots of resources. Therefore, we propose a multi-scale prompt generator~(MSPGenerator) to generate auxiliary masks. Then, points and bounding boxes can be produced based on the auxiliary masks. 

In Figure~\ref{fig:arch}(c), we illustrate our proposed MSPGenerator to solve the requirements of extra proper prompts provided. We extract 5 feature maps from different levels of the image encoder.
% every $n$ time with the feature map of the input of transformer blocks with a size of $BD\times H\times W\times C$. 
% \begin{equation}
% \vspace{-0.02cm}
%         n = (\text{ViT\_Depth} \times H~(W)) / imgH~(imgW).
%     \label{equa:ntimes}
% \vspace{-0.02cm}
% \end{equation}
All extracted feature maps are fed to our designed MSPGenerator to generate auxiliary multi-classes masks. The MSPGenerator is a hierarchical structure built by convolutional and transpose convolution layers. Starting with the deepest feature map, it is gradually upsampled to 2x the size and then concatenates with shallower feature maps. Finally, we obtain the auxiliary masks of the same size as the final segmentation.
% with size ${B\times N{\times} D{\times} imgH{\times} imgW}$. 
To alleviate the gradient vanishing and converge quickly, we involve deep supervision in the MSPGenerator by adding supervision loss at different levels. Then, we utilize the auxiliary masks to generate a point, a bounding box, and a mask for each class. 
% Finally, we obtain the point, bounding box, and mask prompts, which are fed into the prompt encoder to generate the sparse embedding, dense embedding, and image positional embedding.

\vspace{-0.4cm}
\subsection{Proposed Prompt Way} 
\vspace{-0.2cm}
After we obtain auxiliary masks that can produce points and bounding boxes, we should consider which combination of points, boxes, and masks is the best way to segment the medical image. 
Therefore, we conduct experiments that use ground truth to generate points, bounding boxes, and masks for each class as candidate prompts to find the best combination shown in Figure~\ref{fig:point-prompt}(a).
Experiments demonstrate that the prompts of points with or without masks fail, as SAM almost segments the entire chest as the liver class~(the purple region). The failed reason is that the liver region in the raw images has weak boundaries and is similar to other regions. When involving bounding boxes, each class can be located in the corrected region though there are errors. The best performance is the prompt for the combination of points, bounding boxes, and masks. Therefore, we chose the combination of points, bounding boxes, and masks as our model's prompt. However, the selection of points can bring about enormous differences in performance. The criterion is that the selected point should be as representative of a specific object as possible and inside the mask. It means that the point should be as central as possible in the masks. In other words, the points farthest from the boundaries should be selected as the point prompt. Figure~\ref{fig:point-prompt}(b) shows the results of different points with or without bounding boxes. There are enormous differences in performance if randomly selecting points in masks. The performance of selecting the central point of the bounding boxes is not the best and the central point is not always located on masks, such as the Myo class~(green region) in Figure~\ref{fig:point-prompt}(d). When we use the Euclidean distance transform to calculate the distance from the boundaries for each pixel and obtain the candidate points farthest from the boundaries as the point prompt, the performance is the best, which is shown at the right-bottom of Figure~\ref{fig:point-prompt}(b).

\vspace{-0.4cm}
\subsection{Modified Mask Decoder}
\vspace{-0.2cm}
Figure~\ref{fig:arch}(d) illustrates our proposed mask decoder. Since there is an image positional embedding for the mask decoder, we add a learnable depth positional embedding block with image positional embedding to learn extra depth information. The mask decoder consists of two transformers and each transformer consists of two cross-attention blocks, which we also insert DFusedAdapter to do adaption and learn extra depth information. Finally, we would obtain a series of binary segmentation masks for each class.
% , and each output for a certain class has a different distribution from other outputs. 
To properly process multi-class segmentation, we equip our proposed M(ulti)C(lasses)-Adapter which is an invert-bottleneck structure that contains several convolutional layers with a softmax function to adapt binary segmentation to multi-class segmentation. 
% During training, the loss is calculated between the output of MC-Adapter and ground truth.

\noindent \textbf{Proposed MC-Adapter.} The original SAM segments all possible objects by binary masks and does not classify each object belonging to which class, which may result in some pixels being considered belonging to more than 2 classes or not belonging to any class.
% which is more like for instance segmentation methods. 
Our expectation of SAM is to predict each pixel to one specific class given raw images as input, like traditional medical image segmentation. After we obtain all binary masks for the total classes, we observe each binary mask for a certain class has a different distribution from other outputs, since each output is generated by a specific prompt and trained by a sigmoid function. Therefore, it will obtain very bad results if we directly use a softmax function for all output, which is shown in Figure~\ref{fig:point-prompt}(c). We show two different outputs of SAM for the background and liver class. When we individually consider each output, both of the areas of the red boxes are not considered to belong to its class since the intensities of all pixels are smaller than 0, in the range of [-2, -1) and [-1, 0) for background class and liver class, respectively.  However, the area of the red box is considered as the liver class finally when we adopt a softmax function for all outputs since the output intensity for the liver class is the largest but less than 0. Therefore, to adapt the difference and classify each pixel to only one class, we also design an invert-bottleneck architecture that consists of two convolutional layers to adapt binary segmentation to multi-class segmentation, named MC-Adapter.

\vspace{-0.4cm}
\section{Experiments}
\label{sec:experiments}
\vspace{-0.2cm}

\begin{table}[!t]\small
    \setlength{\tabcolsep}{3pt}
    \centering
    % \vspace{-0.4cm}
    \resizebox{0.99\linewidth}{!}{ %< auto-adjusts font size to fill line
    \begin{tabular}{@{}l|ccccccccccccccc|c@{}}
    \toprule
    
    Method & Spleen & R.Kd & L.Kd & GB & Eso. & Liver & Stom. & Aorta & IVC  & Panc. & RAG & LAG & Duo. & Blad. &  Pros. & Average \\
    \midrule
    TransBTS \cite{wang2021transbts} & 0.885 & 0.931 & 0.916 & 0.817 & 0.744 & 0.969 & 0.837 & 0.914 & 0.855 & 0.724 & 0.630 & 0.566 & 0.704 & 0.741 & 0.650 & 0.792 \\
    UNETR \cite{hatamizadeh2022unetr} & 0.926 & 0.936 & 0.918 & 0.785 & 0.702 & 0.969 & 0.788 & 0.893 & 0.828 & 0.732 & 0.717 & 0.554 & 0.658 & 0.683 & 0.722 & 0.762 \\
    nnFormer \cite{zhou2021nnformer} & 0.935 & 0.904 & 0.887 & 0.836 & 0.712 & 0.964 & 0.798 & 0.901 & 0.821 & 0.734 & 0.665 & 0.587 & 0.641 & 0.744 & 0.714 & 0.790 \\
    SwinUNETR \cite{hatamizadeh2021swin} & 0.959 & 0.960 & 0.949 & \textbf{0.894} & 0.827 & 0.979 & 0.899 & 0.944 & 0.899 & 0.828 & 0.791 & 0.745 & 0.817 & 0.875 & 0.841 & 0.880 \\
    nn-UNet~\cite{isensee2019automated} & \textbf{0.965} & 0.959 & 0.951 & 0.889 & 0.820 & \textbf{0.980} & 0.890 & 0.948 & 0.901 & 0.821 & 0.785 & 0.739 & 0.806 & 0.869 & 0.839 & 0.878 \\
    % 3D UX-Net \cite{lee20223d} & \textbf{0.970} & 0.967 & 0.961 & \textbf{0.923} & 0.832 & \textbf{0.984} & 0.920 & 0.951 & 0.914 & 0.856 & \textbf{0.825} & 0.739 & \textbf{0.853} & 0.906 & \textbf{0.876} & 0.900 \\
    \hline

    % MaskSAM (Ours)   & 0.963 & \textbf{0.973} & \textbf{0.969} & 0.864 & \textbf{0.875} & 0.982 & \textbf{0.940} & \textbf{0.962} & \textbf{0.922} & \textbf{0.888} & 0.793 & \textbf{0.813} & \textbf{0.850} & \textbf{0.922} & 0.855 & \textbf{0.905}

    MaskSAM (Ours)   & 0.961 & \textbf{0.969} & \textbf{0.965} & 0.856 & \textbf{0.871} & 0.978 & \textbf{0.938} & \textbf{0.959} & \textbf{0.918} & \textbf{0.882} & \textbf{0.790} & \textbf{0.809} & \textbf{0.847} & \textbf{0.916} & \textbf{0.849} & \textbf{0.901}
    \\
    \bottomrule
    \end{tabular}}
    \caption{The comparison of results on the AMOS testing dataset on the leaderboard.  
    % evaluated by Dice Score. Best results are denoted as \textbf{bold}. 
    }% \caption
    \label{tab:amos}
    \vspace{-0.4cm}
\end{table}

\begin{table}[!t]\small
    \setlength{\tabcolsep}{3pt}
    \centering
    \resizebox{0.99\linewidth}{!}{ %< auto-adjusts font size to fill line
    \begin{tabular}{@{}l|c|cccccccc@{}}
    \toprule
    % Method & DSC & Aotra $\uparrow$ & Gallbladder $\uparrow$  & Kidnery(L) $\uparrow$ & Kidnery(R) $\uparrow$ & Liver $\uparrow$ & Pancreas $\uparrow$ & Spleen $\uparrow$ & Stomach  $\uparrow$ 
    Method & DSC$\uparrow$ & Aotra $\uparrow$ & Gallbladder $\uparrow$  & Kidnery(L) $\uparrow$ & Kidnery(R) $\uparrow$ & Liver $\uparrow$ & Pancreas $\uparrow$ & Spleen $\uparrow$ & Stomach  $\uparrow$ \\ 
    \midrule
    % U-Net \cite{ronneberger2015u} & 74.99  & 83.17 & 58.74 & 80.40 & 73.36 & 93.13 & 45.43 & 83.90 & 66.59 \\ 

    TransUNet \cite{chen2021transunet} & 77.48 &  87.23 & 63.16 & 81.87 & 77.02 & 94.08 & 55.86 & 85.08 & 75.62\\
    SwinUNet \cite{cao2021swin} & 79.13  & 85.47 & 66.53 & 83.28 & 79.61 & 94.29 &  56.58 & 90.66 & 76.6\\
    % TransClaw-U-Net \cite{chang2021transclaw} & 78.09 & 85.87 & 61.38 & 84.83 & 79.36 & 94.28 & 57.65 & 87.74 & 73.55\\
    % LeVit-UNet-384s \cite{xu2021levit} & 78.53 & 87.33 & 62.23 & 84.61 & 80.25 & 93.11 & 59.07 & 88.86 & 72.76\\
    % WAD \cite{li2021more} & 80.30 & 87.73 & 69.93 & 83.95 & 79.78 & 93.95 & 61.02 & 88.86 & 77.16\\
    UNETR \cite{hatamizadeh2022unetr}  & 79.56 & 89.99 & 60.56 & 85.66 & 84.80 & 94.46 & 59.25 & 87.81 & 73.99\\
    nnUNet \cite{isensee2019automated} & 86.21 & 92.39 & 71.71 & 86.07 & \textbf{91.46} & 95.84 & 82.92 & 90.31 & 79.01\\
    nnFormer \cite{zhou2021nnformer} & 86.57 & \textbf{92.40} & 70.17 & 86.57 & 86.25 & 96.84 & \textbf{83.35} & 90.51 & 86.83\\ \hline
    SAMed \cite{zhang2023customized} & 81.88 & 87.77 & 69.11 & 80.45 & 79.95 & 94.80 & 72.17 & 88.72 & 82.06 \\ 
    
    SAMed\_s \cite{zhang2023customized} & 77.78 & 83.62 & 57.11 & 79.63 & 78.92 & 93.98 &  65.66 & 85.81 & 77.49  \\ 
    
    SAM3D \cite{bui2023sam3d} & 79.56 & 89.57 & 49.81 & 86.31 & 85.64 & 95.42 & 69.32 & 84.29 & 76.11 \\ 
    \hline

    Self-Prompt-SAM (Ours)  & \textbf{86.74}  & 91.99 & 69.95 & 85.65 & 85.40 & \textbf{97.39} & 79.18 & \textbf{94.38} & \textbf{89.94}
    \\
    \bottomrule
    \end{tabular}}
    \caption{
    Quantitative results on Synapse dataset (DSC in \%). }
    % Best results are denoted as \textbf{bold}.
    % \bd{Is self-prompt-SAM MaskSAM?}
    % } % \caption
    \label{tab:sotaSynapse}
    \vspace{-1.0cm}
\end{table}

\noindent\textbf{Datasets and Evaluation Metrics.}
We use three publicly available datasets in our experiments, AMOS22 Abdominal CT Organ Segmentation~\cite{ji2022amos}, Synapse multiorgan segmentation~(Synapse)~\cite{landman2015miccai}, and Automatic Cardiac Diagnosis Challenge (ACDC)~\cite{bernard2018deep}.
\textbf{(i)} AMOS22 dataset consists of 200 cases of abdominal CT scans with 16 anatomies manually annotated for abdominal multi-organ segmentation. and we evaluate 200 test images using our model on the AMOS22 leaderboard. 
\textbf{(ii)} Synapse dataset consists of 30 cases of abdominal CT scans. Following the split strategies \cite{chen2021transunet},  we use a random split of 18 training cases and 12 cases for validation. We evaluate model performance via the average Dice score~(DSC) on 8 abdominal organs.
\textbf{(iii)} ACDC dataset consists of 100 patients with labels on the right ventricle (RV), myocardium (MYO) and left ventricle (LV). We use a random split of 70 training cases, 10 validation cases, and 20 testing cases. We evaluate performance by the average DSC.

% \noindent\textbf{Deep Supervision.} Our network is trained with deep supervision when training. Auxiliary losses are added in the decoder to the last three stages~(the three largest resolutions). For each deep supervision output, we downsample the ground truth segmentation mask for the loss computation with each deep supervision output. The final training objective is the sum of all resolutions loss:
% \begin{equation}
%     \begin{aligned}
%             \mathcal{L} = w_1 \cdot \mathcal{L}_1 + w_2 \cdot \mathcal{L}_2 + w_3 \cdot \mathcal{L}_3
%     \end{aligned}
%     \label{equa:finalloss}
% \end{equation}
% where the weights halve with each decrease in resolution~(\textit{i.e.,} $w_2 = \frac{1}{2} \cdot w_1; w_3 = \frac{1}{4} \cdot w_1$, etc), and all weight are normalized to sum to 1. Meanwhile, the resolution of $\mathcal{L}_1$ is equal to $2 \cdot \mathcal{L}_2$ and $4 \cdot \mathcal{L}_3$

\vspace{-0.4cm}
\subsection{Comparison with State-of-the-Art Methods}
\vspace{-0.2cm}
\noindent \textbf{Results on AMOS22.} We compare Self-Prompt-SAM with the methods that are widely used and well-recognized in the community, including the convolution-based method (nnUNet~\cite{isensee2019automated}), transformer-based methods (UNETR~\cite{hatamizadeh2022unetr}, SwinUNETR~\cite{hatamizadeh2021swin}, and nnFormer~\cite{zhou2021nnformer}). To fair comparison, all results are based on 5-fold cross-validation without any ensembles. Table \ref{tab:amos} shows that Self-Prompt-SAM outperforms all existing methods in most organs, achieving a new SOTA performance in DSC. Specifically, it surpasses nnUNet and SwinUNETR by 2.3\% and 2.1\% in DSC, respectively, confirming the efficacy of our method. 

\begin{figure*}[!t]
\centering
    % \vspace{-0.2cm}
\includegraphics[width=1.0\linewidth]{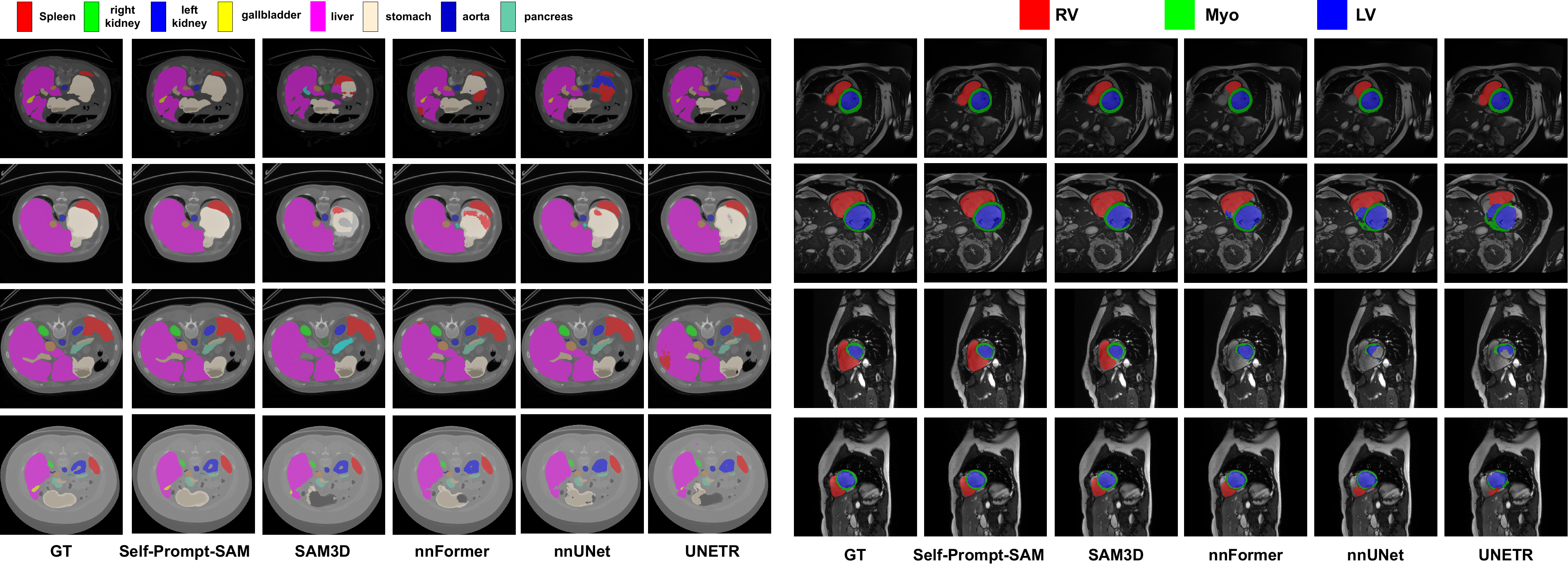}
%\vspace{-0.4cm}
  \caption{Qualitative comparison on the Synapse and ACDC dataset.}
  \vspace{-0.7cm}
\label{fig:sota}
\end{figure*}

\noindent \textbf{Results on Synapse.}~We compare Self-Prompt-SAM with several leading SAM-based method(\textit{i.e.},~SAMed~\cite{zhang2023customized} and SAM3D~\cite{bui2023sam3d}), convolution-based methods (\textit{i.e.}, nnUNet~\cite{isensee2019automated}) and transformer-based methods (\textit{i.e.}, nnFormer~\cite{zhou2021nnformer}). Table~\ref{tab:sotaSynapse} shows that Self-Prompt-SAM outperforms all existing methods and achieves a new SOTA performance. Specifically, our model surpasses SAMed, nnUNet, and nnFormer by $4.9\%$, $0.5\%$, $0.0\%$, and $0.7\%$ in DSC.  Meanwhile, our model predicts well on the large-size labels, `Liver', `Spleen', and `Stomach', due to our proposed DFusedAdapter can learn more 3D spatial information and adapt 2D SAM to medical image segmentation. Fig.~\ref{fig:sota} demonstrates that Self-Prompt-SAM can predict more accurately the `Liver', `Spleen', and `Stomach' labels, demonstrating the effectiveness of our method.

% \textcolor{red}{Explain why it is good at the labels.}
%Therefore, we confirm that our U-MLP method outperforms the other methods. 

% \begin{table}[tp!]
% \centering
% % \vspace{-0.6cm}
% \resizebox{0.5\linewidth}{!}{ %< auto-adjusts font size to fill line
% \begin{tabular}{@{}l|c|ccc@{}}
% \toprule
% Method & Average $\uparrow$ & RV $\uparrow$ & Myo $\uparrow$ & LV $\uparrow$ \\ \midrule
% R50-U-Net \cite{ronneberger2015u} & 87.55 & 87.10 & 80.63 & 94.92\\
% % R50-AttnUNet\cite{schlemper2019attention} & 86.75 & 87.58 & 79.20 & 93.47 \\ 
% VIT-CUP \cite{dosovitskiy2020image} & 81.45 & 81.46 & 70.71 & 92.18 \\
% R50-VIT-CUP \cite{dosovitskiy2020image} & 87.57 & 86.07 & 81.88 & 94.75 \\ 
% UNETR \cite{hatamizadeh2022unetr}  & 88.61 & 85.29 & 86.52 & 94.02\\
% TransUNet \cite{chen2021transunet} & 89.71 & 88.86 & 84.54 & 95.73\\
% SwinUNet \cite{cao2021swin} & 90.00 & 88.55 & 85.62 & 95.83 \\ 
% LeViT-UNet-384s \cite{xu2021levit} & 90.32 & 89.55 & 87.64 & 93.76\\
% nnUNet \cite{isensee2019automated} & 91.61 & 90.24 & 89.24 & 95.36\\
% nnFormer \cite{zhou2021nnformer} & 92.06 & 90.94 & 89.58 & 95.65\\
% SAM3D \cite{bui2023sam3d} & 90.41 & 89.44 & 87.12 & 94.67\\
% \hline
% Self-Prompt-SAM (Ours) & \textbf{93.26} & \textbf{92.20} & \textbf{91.22} & \textbf{96.36} \\
% %Ours & \textbf{91.59} & \textbf{90.17} & \textbf{89.14} & 95.45 \\
% \bottomrule
% \end{tabular}} 
% \caption{Quantitative evaluation with SOTA methods on the ACDC dataset (dice score in \%).}
% \vspace{-0.6cm}
% \label{tab:sotaACDC}
% \end{table}

\begin{wraptable}{r}{5.0cm}
\centering
 	\vspace{-0.6cm}
          \resizebox{1\linewidth}{!}{% 
\begin{tabular}{@{}l|c|ccc@{}}
\toprule
Method & Average $\uparrow$ & RV $\uparrow$ & Myo $\uparrow$ & LV $\uparrow$ \\ \midrule
R50-U-Net \cite{ronneberger2015u} & 87.55 & 87.10 & 80.63 & 94.92\\
% R50-AttnUNet\cite{schlemper2019attention} & 86.75 & 87.58 & 79.20 & 93.47 \\ 
VIT-CUP \cite{dosovitskiy2020image} & 81.45 & 81.46 & 70.71 & 92.18 \\
R50-VIT-CUP \cite{dosovitskiy2020image} & 87.57 & 86.07 & 81.88 & 94.75 \\ 
UNETR \cite{hatamizadeh2022unetr}  & 88.61 & 85.29 & 86.52 & 94.02\\
TransUNet \cite{chen2021transunet} & 89.71 & 88.86 & 84.54 & 95.73\\
SwinUNet \cite{cao2021swin} & 90.00 & 88.55 & 85.62 & 95.83 \\ 
LeViT-UNet-384s \cite{xu2021levit} & 90.32 & 89.55 & 87.64 & 93.76\\
nnUNet \cite{isensee2019automated} & 91.61 & 90.24 & 89.24 & 95.36\\
nnFormer \cite{zhou2021nnformer} & 92.06 & 90.94 & 89.58 & 95.65\\
SAM3D \cite{bui2023sam3d} & 90.41 & 89.44 & 87.12 & 94.67\\
\hline
Self-Prompt-SAM (Ours) & \textbf{93.26} & \textbf{92.20} & \textbf{91.22} & \textbf{96.36} \\
%Ours & \textbf{91.59} & \textbf{90.17} & \textbf{89.14} & 95.45 \\
\bottomrule
\end{tabular}}
\caption{Quantitative evaluation on ACDC (dice score in \%).}
	\vspace{-0.8cm}
 \label{tab:sotaACDC}
\end{wraptable}

\noindent\textbf{Results on ACDC.}~In Table~\ref{tab:sotaACDC}, we compare Self-Prompt-SAM with several leading SAM-based method (\textit{i.e.} SAM3D~\cite{bui2023sam3d}), convolution-based method (\textit{i.e.}, nnUNet~\cite{isensee2019automated}) and transformer-based method (\textit{i.e.}, nnFormer~\cite{zhou2021nnformer}). The results show that Self-Prompt-SAM outperforms various state-of-the-art approaches, surpassing SAM3D, nnUNet, nnFormer by $2.8\%$, $1.2\%$, and $1.6\%$ in DSC, respectively. Fig.~\ref{fig:sota} shows that Self-Prompt-SAM can predict more accurately on all labels. The results demonstrate the effectiveness of our method since our proposed modules can properly solve the drawbacks of SAM when adapting to medical segmentation. %especially in the RV label. Therefore, we confirm that our U-MLP method outperforms other leading methods,  demonstrating the effectiveness of our method.

\vspace{-0.4cm}
\subsection{Ablation Study}
\vspace{-0.2cm}
\label{sec:ablation}
\noindent\textbf{Baseline Models.} The proposed Self-Prompt-SAM has 9 baselines, as shown in Table~\ref{tab:archAblation}. All baselines adopt the whole structure of SAM and only add blocks. (i) S1 adopts a series of stacked CNNs for the prompt generator combined with the image encoder. (ii) S2 utilizes our proposed MSPGenerator to combine with the image encoder to generate prompts. (iii) S3 adds the vanilla adapter~\cite{houlsby2019parameter} with each transformer block in the image encoder and the mask decoder based on S2. (iv) S4 adds the modified adapter by inserting the invert-bottleneck depth MLPs before the adapter with a skip connection based on S2. (v) S5 adds the modified adapter by inserting the invert-bottleneck depth MLPs with a skip connection after the adapter based on S2. (vi) S6 adds our DFusedAdapter with each transformer block in the image encoder and mask decoder based on S2. (vii) S7 adds depth positional embedding~(DPosEmbed) in the image encoder and the mask encoder based on S6. (viii) S8 adds an MAdapter before the image encoder based on S7. (ix) S9 is our full model in Fig.~\ref{fig:arch}, which adds an MC-Adapter to adapt binary segmentation to multi-classes segmentation based on S8. 
% (viii) S8 is our full model, named Self-Prompt-SAM, illustrated in Figure~\ref{fig:arch}. 
% The results of the ablation study are shown in Table~\ref{tab:archAblation}.

% \begin{table}[t!]
%     %\vspace{-0.6cm}
%     \centering
%     \resizebox{0.7\linewidth}{!}{ %< auto-adjusts font size to fill line
%     \begin{tabular}{@{}cl|c}
%     \toprule
%     & Method & DSC $\uparrow$   \\ \midrule
%     S1 & SAM + stacked CNNs prompt generator & 79.57 \\
%     S2 & SAM + MSPGenerator & 82.20   \\ 
%     S3 & SAM + MSPGenerator + vAdapter & 90.08   \\
%     S4 & SAM + MSPGenerator + vAdapter w/ Depth MLPs before vApdater & 91.45   \\ 
%     S5 & SAM + MSPGenerator + vAdapter w/ Depth MLPs after vApdater & 91.52   \\ 
%     S6 & SAM + MSPGenerator + DFusedAdapter & 91.73   \\ 
%     S7 & SAM + MSPGenerator + DFusedAdapter + DPosEmbed & 91.88  \\ 
%     S8 & SAM + MSPGenerator + DFusedAdapter + DPosEmbed + MAdapter & 92.20  \\ \hline
%     % S7 & SAM + MSPGenerator + DFusedAdapter + DPosEmbed + MAdapter + MC-Adapter & 93.04 & 90.07 \\ \hline
%     % S6 & LE-Em+hi-dwLMLP+skips & 91.49 & 90.07 & 88.98 & 95.41 \\ \hline
%     S9 & Our Full Model~(S6 + MC-Adapter) & \textbf{93.26}  \\
%     \bottomrule
%     \end{tabular}} 
%     \caption{The ablation studies of the proposed method on the ACDC dataset. The MSPGenerator means the multi-scale prompt generator. The vAdapter means the vanilla adapter. The DPosEmbed means the depth positional embedding.}
%     \label{tab:archAblation}
%     \vspace{-0.6cm}
% \end{table}

\begin{wraptable}{r}{7.0cm}
\centering
 	\vspace{-0.6cm}
          \resizebox{1\linewidth}{!}{% 

\begin{tabular}{@{}cl|c}
    \toprule
    & Method & DSC $\uparrow$   \\ \midrule
    S1 & SAM + stacked CNNs prompt generator & 79.57 \\
    S2 & SAM + MSPGenerator & 82.20   \\ 
    S3 & SAM + MSPGenerator + vAdapter & 90.08   \\
    S4 & SAM + MSPGenerator + vAdapter w/ Depth MLPs before vApdater & 91.45   \\ 
    S5 & SAM + MSPGenerator + vAdapter w/ Depth MLPs after vApdater & 91.52   \\ 
    S6 & SAM + MSPGenerator + DFusedAdapter & 91.73   \\ 
    S7 & SAM + MSPGenerator + DFusedAdapter + DPosEmbed & 91.88  \\ 
    S8 & SAM + MSPGenerator + DFusedAdapter + DPosEmbed + MAdapter & 92.20  \\ \hline
    % S7 & SAM + MSPGenerator + DFusedAdapter + DPosEmbed + MAdapter + MC-Adapter & 93.04 & 90.07 \\ \hline
    % S6 & LE-Em+hi-dwLMLP+skips & 91.49 & 90.07 & 88.98 & 95.41 \\ \hline
    S9 & Our Full Model~(S8 + MC-Adapter) & \textbf{93.26}  \\
    \bottomrule
    \end{tabular}} 
    \caption{Ablation studies. vAdapter means vanilla adapters. DPosEmbed means depth positional embedding.}
    \vspace{-0.9cm}
    \label{tab:archAblation}
\end{wraptable}

\noindent\textbf{Ablation analysis.} The results are shown in Table~\ref{tab:archAblation} on ACDC.  When we use an MSPGenerator, the DSC of S2 improves by 2.7\% compared to S1 consisting of stacked convolution layers. The result confirms the effectiveness of the proposed MSPGenerator. After inserting vanilla adapters into each transformer block, the performance of S3 is greatly improved 8\% compared to S2 without adapters, demonstrating that using adapters is feasible to fine-tune SAM to medical image segmentation. Meanwhile, we found that the performance of S4 and S5 is very close when we insert the depth MLPs with a skip connection before or after the vanilla adapter. But the DFuserAdapter achieves the best performance compared to S4 and S5. Moreover, S6 improves by 1.7\% compared to S3. The result confirms the effectiveness of DFusedAdapter. When involving depth positional embeddings into the image encoder and mask decoder, the performance of S7 improves by more than 0.1\% compared to S6 without any depth positional embedding, demonstrating the effectiveness of the DPosEmbed. When we adopt the MAdapter before the image encoder, the average DSC of S8 improves by 0.4\% compared to S7, which confirms the benefits of the MAdapter. 
% S9 is our full model, Self-Prompt-SAM, which utilizes an MC-Adapter at the end of SAM to adapt binary segmentation to multiclass segmentation based on S8 as shown in Fig.~\ref{fig:arch}. 
Compared to S8, S9 (our full model, Self-Prompt-SAM) brings 1\% improvements. Therefore, the results demonstrate the effectiveness of Self-Prompt-SAM.

\vspace{-0.4cm}
\section{Conclusion}
\vspace{-0.2cm}
\label{sec:conclusion}
% In this paper, we propose a self-prompt SAM adaptation framework for medical image segmentation, named Self-Prompt-SAM, which adapts pre-trained SAM models worked on from 2D natural images to 3D medical images without any prompts provided. By designing a multi-scale prompt generator~(MSPGenerator) combined with the image encoder to generate auxiliary masks, our model can generate prompts by itself. The auxiliary masks would be used to generate bounding boxes as box prompts and utilize Distance Transform to select the points farthest from the boundary as point prompts. We designed a 3D depth-fused adapter~(DfusedAdapter) and injected DFusedAdapters into each transformer block in the image encoder and mask decoder to enable pre-trained 2D SAM models to extract 3D information and adapt to medical images. 
% Extensive experiments demonstrate that our method achieves state-of-the-art performance and outperforms nnUNet by 2.3\% on AMOS2022~\cite{ji2022amos}, 1.6\% on ACDC~\cite{bernard2018deep} and 0.5\% on Synapse~\cite{landman2015miccai} datasets.

We introduce Self-Prompt-SAM, a framework for adapting pre-trained SAM models from 2D natural images to 3D medical images without manual prompts. Our method employs a multi-scale prompt generator (MSPGenerator) to generate prompts autonomously. These prompts are utilized for bounding boxes and point selection using Distance Transform. We integrate a 3D depth-fused adapter (DfusedAdapter) into the image encoder and mask decoder to enable pre-trained 2D SAM models to process 3D medical images. Extensive experiments show that our method achieves state-of-the-art performance, surpassing nnUNet by 2.3\% on AMOS22, 1.6\% on ACDC, and 0.5\% on Synapse datasets.

\clearpage

\bibliographystyle{splncs04}
\bibliography{miccai}

\clearpage
\appendix

\setcounter{page}{1}
\setcounter{table}{0}
\setcounter{figure}{0}
\setcounter{equation}{0}

\section{Rethinking SAM.}
SAM is the first prompt-driven foundation model for natural image segmentation, which is trained on the large-scale SA-1B dataset of 1B masks and 11M images, allowing the model to have a strong zero-shot generalization. 
SAM consists of three main components, the image encoder that employs the Vision Transformer as the backbone to extract image features, the prompt encoder that embeds various types of prompts, including points, boxes, or texts, and the lightweight mask decoder to generate masks based on the image embedding, prompt embedding, image positional embedding, and output token. 
% To retain all zero-shot capabilities, we keep all structures, freeze all weights, and only add learnable blocks to SAM to perform adaptation. Figure~\ref{fig:arch} illustrates our proposed Self-Prompt-SAM, which consists of a modified image encoder, a designed MSPGenerator, the original prompt encoder, and a modified mask decoder.

There are some issues with adapting SAM to medical segmentation tasks. i) SAM needs users to provide appropriate prompts, such as points or boxes, to locate the target regions. We cannot expect users to have a medical background; therefore, we designed an MSPGenerator to handle this issue. We introduce it in the following section. ii) SAM does not provide any semantic information for segmentation results since it only generates a binary mask for one prompt. we utilize the MSPGenerator to generate auxiliary multi-class masks, which can be encoded to one-hot binary masks whose number channel dimension is the same number of classes. Therefore, each channel-wise binary mask is specifically responsible for a certain semantic label by the location or index of the channel dimension in the one-hot binary masks. 
% which can produce box prompts and point prompts. 
Sometimes, channel-wise binary masks may not have any foreground. In this case, we assign the values of both a box prompt and a point prompt to zero. 
% In this way, we utilize the location in the channel dimension of the one-hot binary masks to represent the semantic labels. 
iii) The performance of SAM in medical image segmentation usually does not meet the strict requirements in clinical medicine. Therefore, we designed DFusedAdapter, inserted learnable positional embeddings into the image encoder and mask decoder, and designed other blocks for special functionalities to improve the performance to fine-tune.

\section{Theoretical comparison with SAM-based models.} 
The main contributions of our Self-Prompt-SAM different from the existing SAM-based models are i) \underline{automatic prompt}, ii) to provide \underline{semantic labels} for each mask, and iii) \underline{remain all parameters} of the original SAM for zero-shot capabilities. There are several categories of the existing SAM-based models. The first category does not modify SAM, such as MedSAM and Polyp-SAM. These models need manual prompts, such as points or boxes, and cannot classify masks into semantic labels. 
% Meanwhile, reference [r1] adopts U-Net to produce masks instead of manual prompts and just uses SAM without any modification. [r1] cannot classify masks into semantic labels and directly using SAM for medical image segmentation results in subpar performance.
% and exhibits imperfections or total failures in more challenging situations. 
The second category uses parameter-efficient transfer learning, such as Adapters, into SAM. The popular model, Med-SA, uses the GT to generate prompts during inference, which do not have any practical clinical values. It also includes the non-automatic models of the 3DSAM-Adapter and MA-SAM. These models do not handle the requirements of extra prompts. 
% Therefore, our model proposes a prompt generator module to solve this problem. 
The third category is that cannot provide semantic labels to binary masks, such as DeSAM, Med-SA, and MA-SAM. Since SAM only predicts binary masks, these models do not address the lack of representation of semantic labels. The fourth category is abandoning the components of SAM, such as Mask Decoder, to handle the inability to classify semantic labels, such as 3DSAM-Adapter. This way inevitably destroys the consistency and zero-shot capabilities of SAM. These models only use the pre-trained ViT encoder, which is not the contribution of SAM. 

\section{More details for DFusedAdapter}
The DFusedAdapter can be expressed as $\text{DFusedAdapter}(\textbf{\textit{X}}) =$ 
\begin{equation}
    \textbf{\textit{X}} + (\sigma(\textbf{\textit{X}}\cdot \textbf{\textit{W}}_{dn}) + \sigma (\sigma(\textbf{\textit{X}}\cdot \textbf{\textit{W}}_{dn}) \cdot \textbf{\textit{W}}_{Dup}) \cdot \textbf{\textit{W}}_{Ddn}) \cdot \textbf{\textit{W}}_{up},
\label{equa:DFusedAdapter}
\end{equation}
where $\sigma$ denotes the activation function, $\textbf{\textit{W}}_{dn} {\in} \mathbb{R}^{C\times \frac{C}{4}}$ and $\textbf{\textit{W}}_{up} {\in} \mathbb{R}^{\frac{C}{4}\times C}$ denote the linear down- and up-projection layer processing on the channel dimension respectively, $\textbf{\textit{W}}_{Dup} {\in} \mathbb{R}^{D\times 4D}$ and $\textbf{\textit{W}}_{Ddn} {\in} \mathbb{R}^{{4D}\times D}$ denote the linear up- and down-projection layer processing on the depth dimension respectively. 
In this way, our model can learn the extra-depth information.

\section{Implementation Details}
We utilize some data augmentations such as rotation, scaling, Gaussian noise, Gaussian blur, brightness, and contrast adjustment, simulation of low resolution, gamma augmentation, and mirroring. We set the initial learning rate to 0.01 and employ a ``poly'' decay strategy in Eq.~\eqref{equa:polydecay}.
\begin{equation}
    lr(e)= init\_lr \times (1 - \frac{e}{\rm MAX\_EPOCH})^{0.9},
\label{equa:polydecay}
\end{equation}
where $e$ means the number of epochs, MAX\_EPOCH means the maximum of epochs, set it to 1000 and each epoch includes 250 iterations. We utilize SGD as our optimizer and set the momentum to 0.99. The weighted decay is set to 3e-5. We utilize both cross-entropy loss and dice loss by simply summing them up as the loss function. We utilize instance normalization as our normalization layer. Since we expect relatively good auxiliary masks to finetune the mask decoder, only the deep supervision loss of MSPGenerator is trained in the first two hundred epochs. After two hundred epochs, our model combines the deep supervision loss of MSPGenerator and the loss of MC-Adapter at the end of the mask decoder. All experiments are conducted using two NVIDIA RTX A6000 GPUs with 40GB memory.

\noindent\textbf{Deep Supervision.} Our network is trained with deep supervision when training for the auxiliary losses. Auxiliary losses are added in the decoder.
% to the last three stages~(the three largest resolutions). 
For each deep supervision output, we downsample the ground truth segmentation mask for the loss computation with each deep supervision output. The final training objective is the sum of all resolutions loss:
\begin{equation}
    \begin{aligned}
            \mathcal{L} = w_1 \cdot \mathcal{L}_1 + w_2 \cdot \mathcal{L}_2 + w_3 \cdot \mathcal{L}_3 + \cdot \cdot \cdot w_n \cdot \mathcal{L}_n
    \end{aligned}
    \label{equa:finalloss}
\end{equation}
where the weights halve with each decrease in resolution~(\textit{i.e.,} $w_2 = \frac{1}{2} \cdot w_1; w_3 = \frac{1}{4} \cdot w_1$, etc), and all weight are normalized to sum to 1. 
% Since our model has a varied number of stages in the decoder, the number of $n$ is based on the number of stages in the decoder. 
Meanwhile, the resolution of $\mathcal{L}_1$ is equal to $2 \cdot \mathcal{L}_2$ and $4 \cdot \mathcal{L}_3$.

\end{document}